\pdfoutput=1
\documentclass[11pt]{article}

\usepackage{acl}

\usepackage{times}
\usepackage{latexsym}

\usepackage[T1]{fontenc}
\usepackage[utf8]{inputenc}

\usepackage{microtype}

\usepackage{inconsolata}
\usepackage{url}
\usepackage{booktabs}
\usepackage{multirow}
\usepackage{graphicx}
\usepackage{amsfonts}
\usepackage{amssymb}
\usepackage{bbding}
\usepackage{dashrule}
\usepackage{subfigure}
\usepackage{colortbl}
\usepackage{amsmath}

\title{\centering RoTBench: A Multi-Level Benchmark for Evaluating the Robustness of Large Language Models in Tool Learning}

\author{
    \bf{\normalsize
    Junjie Ye$^{1}$,\ \ Yilong Wu$^{1}$,\ \ Songyang Gao$^{1}$,\ \ Caishuang Huang$^{1}$,\ \ Sixian Li$^{1}$,} \\
    \bf{\normalsize \ \ Guanyu Li$^{1}$,\ \ }
    \bf{\normalsize Xiaoran Fan$^{1}$,\ \ }
    \bf{\normalsize Qi Zhang$^{1,3}$\thanks{Corresponding authors.},\ \ Tao Gui$^{2,3*}$,\ \ Xuanjing Huang$^{1,3}$} \\ 
  {$^1$ \normalsize School of Computer Science, Fudan University} \\
  {$^2$ \normalsize Institute of Modern Languages and Linguistics, Fudan University} \\
  {$^3$ \normalsize Shanghai Key Laboratory of Intelligent Information Processing, Fudan University}\\
  \texttt{\normalsize jjye23@m.fudan.edu.cn}\\
  \texttt{\normalsize \{qz, tgui\}@fudan.edu.cn} \\
  }

\begin{document}
\maketitle
\begin{abstract}
Tool learning has generated widespread interest as a vital means of interaction between Large Language Models (LLMs) and the physical world. Current research predominantly emphasizes LLMs' capacity to utilize tools in well-structured environments while overlooking their stability when confronted with the inevitable noise of the real world.
To bridge this gap, we introduce \emph{RoTBench}, a multi-level benchmark for evaluating the robustness of LLMs in tool learning. Specifically, we establish five external environments, each featuring varying levels of noise (i.e., Clean, Slight, Medium, Heavy, and Union), providing an in-depth analysis of the model's resilience across three critical phases: tool selection, parameter identification, and content filling.
Experiments involving six widely-used models underscore the urgent necessity for enhancing the robustness of LLMs in tool learning.
For instance, the performance of GPT-4 even drops significantly from 80.00 to 58.10 when there is no substantial change in manual accuracy.
% , a challenge that persists as their capabilities advance. 
More surprisingly, the noise correction capability inherent in the GPT family paradoxically impedes its adaptability in the face of mild noise.
In light of these findings, we propose RoTTuning, a strategy that enriches the diversity of training environments to bolster the robustness of LLMs in tool learning.
The code and data are available at~\url{https://github.com/Junjie-Ye/RoTBench}.
\end{abstract}

\section{Introduction}

Tool learning has emerged as a critical concept for empowering large language models (LLMs)~\cite{GPT-3, Claude, LLaMA} to interact with the real world~\cite{foundation, augmented-survey, tool-learning, ToolSword}. In this context, the external environment of an LLM contains an ensemble of integrated tools. Each tool is uniquely identified by its name and is described by a succinct paragraph that explains its functionality. Similarly, every parameter within these tools is characterized by its name, along with a description that clarifies its purpose, its optionality, and other pertinent details.

\begin{figure}[!t]
    \centering
    \includegraphics[width=0.9\linewidth]{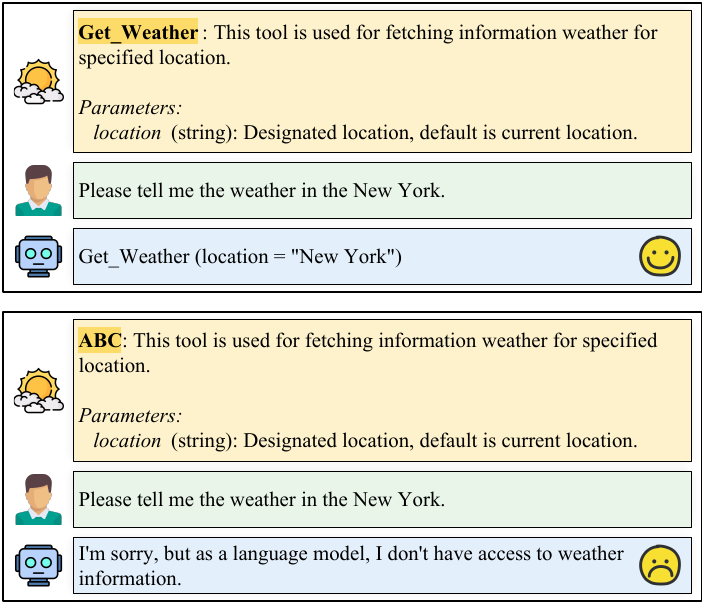}
    \caption{Example of noise affecting tool selection for LLMs. Although the functionality of the tool remains unaffected by its name, renaming ``Get\_Weather'' as ``ABC'' impedes LLMs from utilizing the tool properly.}
    \label{fig:document}
    % \vspace{-4mm}
\end{figure}

Recent research has centered on examining how well LLMs can effectively employ tools within a carefully designed and stable environment. From one perspective, specific studies have scrutinized the outcomes of LLMs' tool usage, verifying both the accuracy of tool selection and the efficacy of the generated responses~\cite{Toolllm, metatool}. This analysis involved evaluating the relevance of the selected tools and the final responses in fulfilling users' requirements. On the other hand, other investigations have delved into the intricate process of tool utilization by LLMs, striving for a more comprehensive assessment of their performance in tool learning~\cite{T-eval, ToolEyes}. This includes an analysis of the diverse capabilities necessary for LLMs to excel in tool learning while also identifying any limitations they may have in this regard.

However, these studies fail to account for the robustness of LLMs in the face of inevitable noise in real-world scenarios~\cite{robust_1, robust_3}. 
Using Figure~\ref{fig:document} as a reference, LLMs recognize the tool for querying weather information when named ``Get\_Weather,'' but not when named ``ABC,'' despite the tool's functionality remaining unaffected by its name.
Consequently, it becomes imperative to investigate whether LLMs can proficiently identify these tools and configure parameters to meet user needs in noisy real-world environments. This research is essential to guarantee their reliability in practical applications.

To fill this gap, we introduce \emph{RoTBench}, a multi-level benchmark for evaluating the robustness of LLMs in tool learning. Specifically, we establish five external environments, which can be categorized as Clean, Slight, Medium, Heavy, and Union in ascending order of noise levels. By evaluating the performance of LLMs across three critical stages: tool selection, parameter identification, and content filling, we aim to offer a thorough and intricate analysis of the stability and reliability of LLMs in tool utilization.

Through experiments conducted on six widely-used LLMs, we observe that the performance of these models is remarkably sensitive to noise. For instance, the performance of GPT-4 even drops significantly from 80.00 to 58.10 when there is no substantial change in manual accuracy. This underscores the pressing requirement to enhance the robustness of LLMs in tool learning. Interestingly, the GPT family of models' inherent noise correction capability appears to hinder its performance in mildly noisy environments.

In light of these findings, we introduce RoTTuning, a technique aimed at augmenting the adaptability of LLMs to a wide range of environments by introducing greater environmental diversity during the training phase. Our experimental results demonstrate that our approach yields an average performance improvement of 16.10 points across diverse environments.

The main contributions of our work are summarized as follows:
\begin{itemize}
    \item We introduce RoTBench, a benchmark designed to evaluate the robustness of LLMs in tool learning. This benchmark contains five environments with different levels of noise, enabling a comprehensive evaluation of robustness throughout three pivotal phases of model tool learning.
    \item The experimental analyses conducted on six widely-used models underscore the imperative of improving the robustness of LLMs in tool learning. These analyses also reveal conflicts between the inherent capabilities of the models and their robustness.
    \item We introduce RoTTuning, a training method for tool learning that focuses on augmenting environmental diversity. Our experiments demonstrate that this approach can effectively enhance LLMs robustness.
\end{itemize}

\section{Related Work}
\paragraph{Analysis of Tool Learning}
Given their extensive world knowledge and superior natural language understanding, researchers have made attempts to leverage LLMs for a wide range of everyday applications~\cite{analy-ye}. In order to push the boundaries of their capabilities, some scholars have proposed enhancing LLMs with external tools, which has gained widespread acceptance~\cite{Toolformer, Toolalpaca}. As research in this area has deepened, certain scholars have summarized the progress made in tool learning for LLMs~\cite{augmented-survey, tool-learning}, sought to uncover developmental insights, and trained more specialized LLMs for tool learning based on these findings~\cite{Toolllm, ToolQA, Toolken}.
Furthermore, recognizing the complexity of tool learning, some researchers have specialized in evaluating not only the outcomes of tool learning~\cite{metatool} but also the entire process~\cite{T-eval, ToolEyes}. However, it's worth noting that all of these current efforts primarily consider LLMs' tool usage in controlled environments, neglecting the inherent complexities of real-life scenarios.
Therefore, we have undertaken an in-depth analysis of the robustness of LLMs in tool learning to advance research in a real-world context.

\begin{figure*}[!t]
    \centering
    \includegraphics[width=\linewidth]{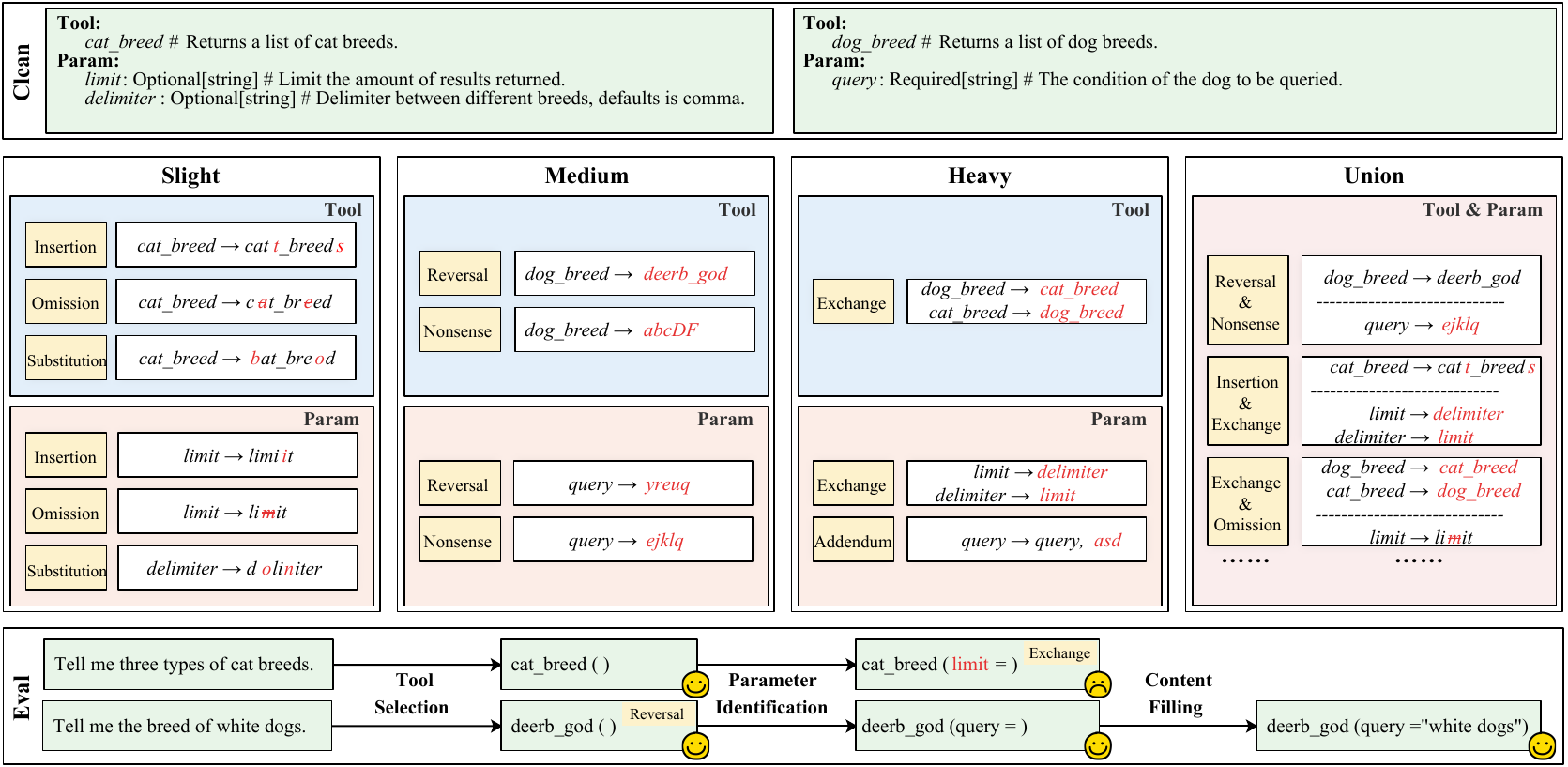}
    \caption{The framework of RoTBench. RoTBench encompasses five environments (i.e., Clean, Slight, Medium, Heavy, and Union), each introduces various noise to the tool and parameters, facilitating a thorough evaluation of the robustness performance of LLMs throughout the three stages of tool usage (i.e., tool selection, parameter identification, and content filling).}
    \label{fig:RoTBench}
    % \vspace{-4mm}
\end{figure*}

\paragraph{Robustness Testing of LLMs}

Robustness is a critical factor in determining the stability of LLMs and plays a pivotal role in their practical deployment in real-life applications, which has garnered significant attention from scholars. In the early stages of research, some scholars conducted tests to assess the robustness of ChatGPT across various natural language processing tasks, highlighting the substantial room for improvement in the current robustness of LLMs~\cite{robust_11,analysis-chen}. Subsequently, other researchers specialized in creating benchmarks, such as PromptBench~\cite{PromptBench}, to examine the consistency of LLM responses by introducing noise into the prompts.
Given that tool learning is poised to extend the capabilities of LLMs and its outcomes can directly impact the state of the physical world~\cite{ToolEyes}, it becomes imperative to thoroughly evaluate its robustness.

\begin{table}[!t]
    \centering
    \resizebox{\linewidth}{!}
    {
    \begin{tabular}{ccccc}
    \toprule
         \textbf{\# Sce}&  \textbf{\# Query} &\textbf{\# Cat} &\textbf{\# Subcat} &\textbf{\# Tool} \\ \midrule
         7& 
     105& 41& 95&568 \\ \bottomrule
     \end{tabular}
     }
    \caption{Statistics information of the data. ``\# Sce'', ``\# Query'', ``\# Cat'', ``\# Subcat'', and ``\# Tool'' correspond to the count of scenarios, user queries, tool categories, tool subcategories, and individual tools, respectively.}
    \label{tab:data}
    % \vspace{-4mm}
\end{table}

\section{RoTBench}
\label{sec:RoTBench}

As depicted in Figure~\ref{fig:RoTBench}, RoTBench encompasses five environments, each characterized by varying levels of noise, facilitating a thorough evaluation of the robustness of LLMs throughout the three stages of tool usage. 
% Detailed descriptions are provided in the following sections.

\subsection{Data Collection}

In order to thoroughly cater to real-world requirements and encompass commonly utilized tools, we utilize ToolEyes~\cite{ToolEyes}, an evaluation system designed for tool learning. This system defines seven real-world application scenarios.
% that span text generation, data understanding, real-time search, application manipulation, personal life, information retrieval, and financial transactions.
Within each of these scenarios, we randomly select 15 user queries for analysis. Since the raw data offers tool information without standardized invocation paths, we have manually labeled these paths to facilitate the evaluation process. Detailed statistics of the data can be found in Table~\ref{tab:data}.

\subsection{Environments Construction}
% In order to thoroughly evaluate the robustness of LLMs in tool learning, and with reference to the hierarchical classification of environmental noise in existing studies, we create five external environments with varying levels of noise: Clean, Slight, Medium, Heavy, and Union. These noise apply to both the tool and its parameters.
To comprehensively assess the resilience of LLMs in tool learning, we reference the hierarchical classification of noise in previous studies~\cite{Textflint, PromptBench, demonsf} and design five distinct external environments. These environments feature varying noise levels that affect both the tool and its parameters.

% \paragraph{Clean}
\textbf{Clean-level} environment employs a runtime framework developed by ToolEyes. This framework furnishes essential information to LLMs for comprehending tools, where the name of each tool epitomizes its functionality and the names of parameters signify their respective meanings. This environment comprises a total of 105 test cases. The remaining four environments are derivatives of this primary environment, each modified by incorporating distinct levels of noise.

% \paragraph{Slight}
\textbf{Slight-level} environment encompasses three types of noise: \emph{insertion}, \emph{omission}, and \emph{substitution}. These correspond to real-world occurrences such as an excess of characters, missing characters, and character errors when naming tools or parameters.
% To minimize the impact of this noise on the overall environmental information
Specifically, we introduce noise in the following ways:
1) We randomly select half of the available tools within the environment. For these selected tools, a random form of noise is applied, altering up to 1/3 of the characters,
% . The remaining tools remain unaltered
resulting in the creation of 105 new data points.
2) For each tool, we randomly select half of the parameters and introduce noise into their names using the method described above,
% . The remaining parameters remain unchanged, 
generating an additional 105 new data entries.
By combining these two approaches, we create a Slight-level environmental test set consisting of 210 test cases.

% \paragraph{Medium}
\textbf{Medium-level} environment introduces two types of noise: \emph{reversal} and \emph{nonsense}. These mirror real-world scenarios where names are reversed or replaced with random strings, rendering the information meaningless.
To apply noise, we follow these procedures:
1) We randomly select half of the available tools. For these tools, there is a 50\% probability that their names will be substituted with random strings, each containing up to 10 characters. Additionally, there is a 50\% chance that the names of these tools will be reversed. This process yields 105 test cases.
2) For each tool, half of the parameters are randomly chosen. These parameters may undergo a 50\% chance of having their names substituted with random strings, each containing up to 5 characters, or a 50\% chance of being reversed. This leads to 105 test cases.
It is worth noting that if the reversal process does not alter the name, it will be replaced with a random string. Consequently, we have successfully generated 210 test cases for the Medium-level environment.

\begin{table*}[!t]
\centering
\resizebox{\linewidth}{!}
{
\begin{tabular}{l | c c |c c |c c| c}
\toprule
\multirow{3}{0.1\linewidth}{\textbf{Models}} & \multicolumn{4}{c|}{\textbf{Open-Source LLMs}} & \multicolumn{2}{c|}{\textbf{Closed-Source LLMs}} & \multirow{3}{0.1\linewidth}{\centering \textbf{\textit{Human}}}\\ \cmidrule(lr){2-5} \cmidrule(lr){6-7}

& \multirow{2}{0.13\linewidth}{\centering \textbf{ToolLLaMA-2-7B-v1}} & \multirow{2}{0.13\linewidth}{\centering \textbf{ToolLLaMA-2-7B-v2}} & \multirow{2}{0.13\linewidth}{\centering \textbf{NexusRaven-13B-v1}} & \multirow{2}{0.13\linewidth}{\centering \textbf{NexusRaven-13B-v2}} & \multirow{2}{0.13\linewidth}{\centering \textbf{GPT-3.5-turbo}} & \multirow{2}{0.13\linewidth}{\centering \textbf{GPT-4}} &  \\
&&&&&&&\\
\midrule
\multicolumn{8}{c}{\textit{Tool Selection}} \\
\midrule
\rowcolor{gray!10} \textbf{Clean} & 66.67 & 70.48 & 55.24 & 73.33 & 75.24 & \textbf{80.00} 
& \textit{88.57}\\
\textbf{Slight} & 57.62 & 65.71 & 52.86 & 76.19 & 59.05 & \textbf{77.14}
& \textit{88.57}\\
\textbf{Medium} & 56.67 & 59.52 & 53.33 & 72.38 & 69.52 & \textbf{84.29}
& \textit{88.57}\\
\textbf{Heavy} & 43.33 & 46.67 & 44.29 & \textbf{62.38} & 56.19 & 60.00 
& \textit{85.71}\\
\textbf{Union} & 44.76 & 43.81 & 42.86 & 56.19 & 53.33 & \textbf{58.10} & \textit{85.71}\\
\midrule
\multicolumn{8}{c}{\textit{Parameter Identification}} \\
\midrule
\rowcolor{gray!10} \textbf{Clean} & 45.71 & 43.81 & 15.24 & \textbf{56.19} & 47.62 & 52.38 
& \textit{88.57}\\
\textbf{Slight} & 40.95 & 40.00 & 17.14 & \textbf{56.67} & 28.10 & 44.29 
& \textit{85.71}\\
\textbf{Medium} & 38.10 & 35.71 & 14.76 & 50.48 & 44.29 & \textbf{53.81} 
& \textit{82.86}\\
\textbf{Heavy} & 28.10 & 27.14 & 10.00 & \textbf{37.62} & 24.29 & 32.86 
& \textit{80.00}\\
\textbf{Union} & 35.24 & 27.62 & 11.43 & 37.14 & 27.62 & \textbf{39.05} & \textit{82.86}\\
\midrule
\multicolumn{8}{c}{\textit{Content Filling}} \\
\midrule
\rowcolor{gray!10} \textbf{Clean} & 28.57 & 25.71 & 1.90 & 37.14 & 30.48 & \textbf{40.00} 
& \textit{74.29}\\
\textbf{Slight} & 24.29 & 23.81 & 3.33 & \textbf{39.05} & 20.00 & 35.71 
& \textit{74.29}\\
\textbf{Medium} & 22.38 & 20.95 & 1.90 & 33.81 & 30.48 & \textbf{46.19} 
& \textit{71.43}\\
\textbf{Heavy} & 14.29 & 14.76 & 0.95 & \textbf{30.00} & 16.19 & 25.24 
& \textit{68.57}\\
\textbf{Union} & 16.19 & 16.19 & 1.90 & 22.86 & 18.10 & \textbf{30.48} & \textit{71.43}\\
\bottomrule
\end{tabular}
}
\caption{Performance of various LLMs in different environments, with the best performance in each environment highlighted in \textbf{bold}. ``Human'' signifies the average level of human performance.}
\label{tab:turn1}
% \vspace{-4mm}
\end{table*}

% \paragraph{Heavy}
\textbf{Heavy-level} environment encompasses two disruptive types of noise: \emph{exchange} and \emph{addendum}, reflecting real-world occurrences of name swapping and information supplementation.
Noise is introduced as follows:
1) All tool names within the environment are randomly shuffled. This shuffling disrupts the association between a tool's name and its functional description, challenging LLMs to accurately comprehend the tool's function despite the disorganized name. This process yields 105 test cases.
2) Half of the tools are randomly chosen, and a new mandatory parameter is introduced with a 50\% probability. This parameter is given a name consisting of a random string of up to 5 characters. LLMs are tasked with providing a specific string of up to 3 characters for the parameter based on its descriptive meaning. The names of these parameters are randomly shuffled with a 50\% probability. For tools with fewer than two parameters, noise is introduced by directly adding new parameters. This process also results in 105 test cases.
In total, 210 Heavy-level environmental test cases have been generated.

% \paragraph{Union}
\textbf{Union-level} environment encompasses all previously mentioned noise categories. Given that the prior noise environments already include noise for both tools and parameters, we randomly choose one noise generation method that impacts tool names and another method that affects parameters from the three previous environment levels. These selected methods are simultaneously applied to generate 105 test cases where both tool names and parameters are subjected to noise injection.

\subsection{Staged Evaluation}
\label{sec:eval}
% The process of LLMs' tool learning contains three crucial stages: tool selection, parameter identification, and content filling.
We evaluate the robustness performance of LLMs at each of stages in tool learning and analyze their respective variations.

% \paragraph{Tool Selection}
\textbf{Tool selection} marks the initial phase of tool usage by LLMs. During this process, LLMs identify suitable tools for addressing the user's query by interpreting the functional descriptions offered by the external environment and subsequently output the names of these tools. It should be emphasized that the name of the tool is essentially a label; the practical deployment of the tool is governed by its functional description.
In evaluating a test case, the score for its tool selection is defined as follows:
\begin{equation}
    s_{TS} = \mathbb{I}(t=\hat{t})
\end{equation}
Here, $\mathbb{I}(x)$ equals 1 if the condition $x$ is true, and 0 otherwise. In this context, $t$ represents the tool chosen by the LLMs, while $\hat{t}$ denotes the tool that needs to be selected.

% \paragraph{Parameter Identification}
\textbf{Parameter identification} involves recognizing the required parameters and outputting their respective names based on their specified needs, following the selection of the appropriate tool. This process necessitates choosing the mandatory parameters, while the optional ones are selected based on actual requirements. Similar to tool selection, the name of the parameter serves as an identifier; however, it is the description of the parameter that truly defines its meaning. 
% Additionally, the sequence in which the parameters are arranged holds no significance.
For each given test case, its parameter identification score is defined as follows:
\begin{equation}
    s_{PI} = s_{TS} \cdot \mathbb{I}(P=\hat{P})
\end{equation}
In this equation, $P$ denotes the set of parameters identified by LLMs, and $\hat{P}$ represents the set of parameters that should be identified.

% \paragraph{Content Filling}
\textbf{Content filling} constitutes the concluding phase in the tool usage process. Once the tool and its corresponding parameters have been selected, LLMs are tasked with breaking down the user-provided information for populating the content of these parameters. Upon accomplishing this step, LLMs formally conclude the entire tool usage cycle, paving the way to receive the tool's output phase and initiate a new interaction.
For each test case, we define a content filling score as follows:
\begin{equation}
    s_{CF} = s_{PI} \cdot \prod_{i=1}^{N}\mathbb{I}(c_i=\hat{c_i})
\end{equation}
Here, $N$ represents the total number of parameters required to be filled. $c_i$ is the content filled by LLMs for the $i$th parameter, and $\hat{c_i}$ refers to the correct content for that parameter.

\section{Experiments}

\subsection{Model Selection}
To evaluate the robustness of widely-used LLMs with tool-use capabilities, we opt for testing four open-source models (i.e., ToolLLaMA-2-7B-v1 \cite{Toolllm}, ToolLLaMA-2-7B-v2 \cite{Toolllm}, NexusRaven-13B-v1 \cite{nexusraven}, NexusRaven-13B-v2 \cite{nexusraven-v2}) and two closed-source models (i.e., GPT-3.5-turbo\footnote{\url{https://platform.openai.com/docs/models/gpt-3-5}}, GPT-4~\cite{GPT-4}).\footnote{The details of LLMs can be found in Appendix~\ref{sec:models}.}

\begin{table}[!t]
    \centering
    \resizebox{\linewidth}{!}{
    \begin{tabular}{p{0.15\linewidth}|l|cc}
    \toprule
     \textbf{Source} & \textbf{Models} & \textbf{F Statistic} & \textbf{P Value} \\ \midrule
   \multirow{4}{0.2\linewidth}{\textbf{Open-Source}}   
    &  \textbf{ToolLLaMA-2-7B-v1} & 2.47 & $4.36 \times 10^{-2}$ \\
    &  \textbf{ToolLLaMA-2-7B-v2} & 3.28 & $1.10 \times 10^{-2}$ \\
    \cmidrule{2-4}
    & \textbf{NexusRaven-13B-v1} & 0.76 & $5.55 \times 10^{-1}$ \\
    & \textbf{NexusRaven-13B-v2} & 6.01 & $9.13 \times 10^{-5}$ \\ \midrule
  \multirow{2}{0.2\linewidth}{\textbf{Closed-Source}} 
    & \textbf{GPT-3.5-turbo} & 6.76 & $2.33 \times 10^{-5}$ \\
    & \textbf{GPT-4} & 5.31 & $3.19 \times 10^{-4}$ \\
  \midrule
\textbf{\textit{Human}} & \textbf{--} & \textit{0.04}& $\mathit{1.00}$\\
 \bottomrule
    \end{tabular}
    }
    \caption{Welch's ANOVA for $s_{CF}$ across the five enviroments for various LLMs.  A p-value below 0.05 indicate significant differences in the data.}
    \label{tab:Welch}
    % \vspace{-4mm}
\end{table}

\subsection{Main Results}

As tool learning involves multiple turns of interaction between LLMs and the environment~\cite{tool-learning,ToolEyes}, with intricate intermediate trajectories that cannot be easily compared, our emphasis lies on evaluating the robustness 
% of their performance in 
% a single turn of interaction. Specifically, we evaluate 
% the performance 
of various LLMs during their initial use of the tool and present the results in Table~\ref{tab:turn1}.\footnote{The results presented are averages across various scenarios, with specific outcomes for each scenario detailed in Appendix~\ref{sec:scenarios}.}
The resulting data reveals intriguing observations.

\textbf{The robustness of current LLMs in tool learning presents considerable scope for enhancement.}
While human performance remains relatively stable across different environments, the performance of LLMs exhibits significant fluctuations. For instance, when transitioning from Clean-level environment to Union-level, human performance in tool selection only decreases by 2.86 points, whereas the average performance of all LLMs decreases by approximately 20.32 points.
To gain a clearer understanding, we employ Welch's ANOVA~\cite{Welch} to analyze the significance of LLMs' performance during the content-filling stage across various environments. As illustrated in Table~\ref{tab:Welch}, our findings underscore the consistency of human performance and the noteworthy disparities in LLMs' performance across different environments.
Consequently, enhancing the robustness of LLMs in tool learning is an area that requires significant attention.

\begin{figure}[!t]
    \centering
    \includegraphics[width=\linewidth]{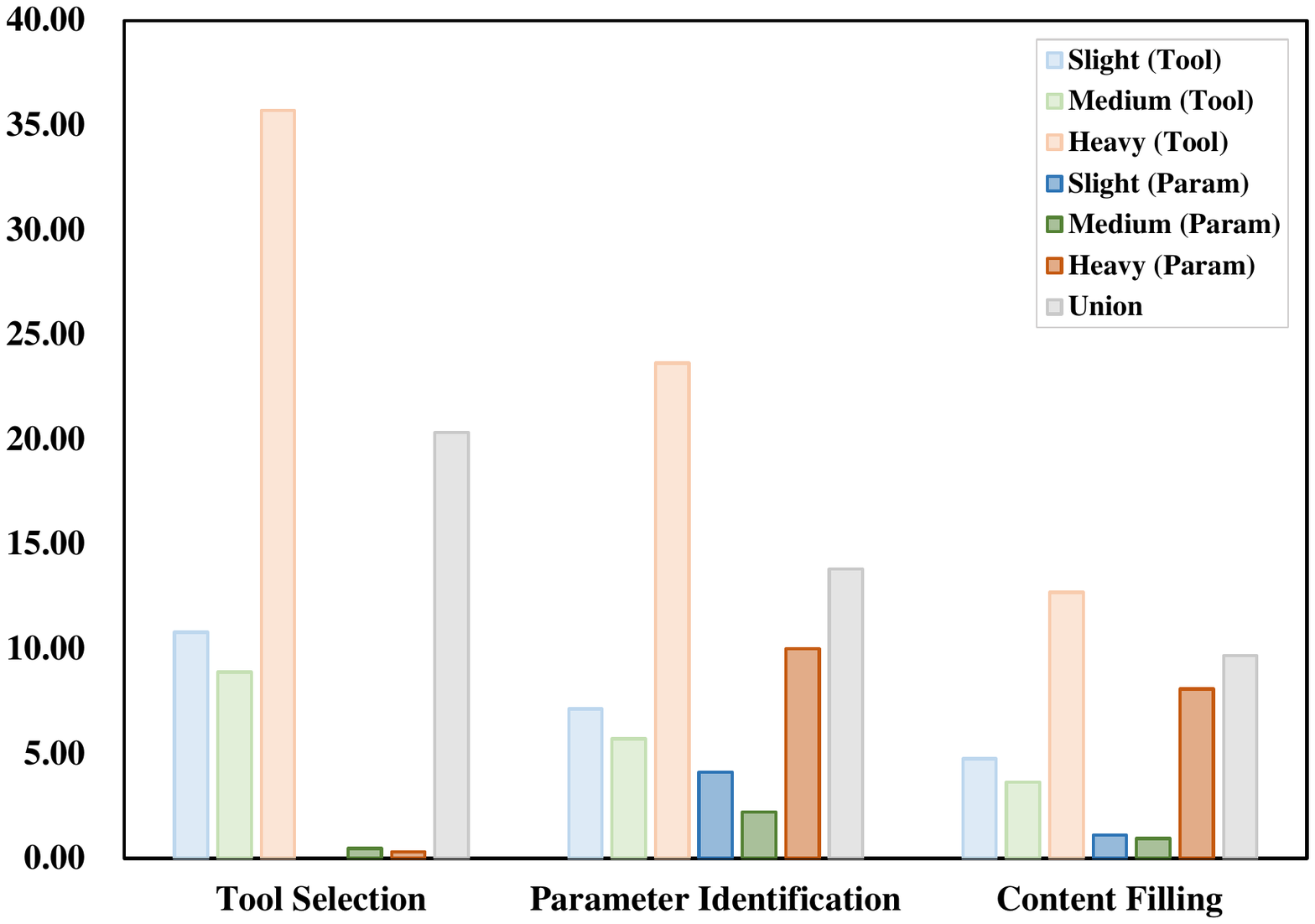}
    \caption{Absolute difference between the average performance of LLMs in various noisy environments and their average performance in Clean-level environment.}
    \label{fig:noise}
    % \vspace{-4mm}
\end{figure}

\textbf{Noise affecting tool names has a more pronounced impact on LLM performance than noise introduced to parameters. }
% As detailed in Section~\ref{sec:RoTBench}, except for the Union-level environment, where noise is introduced to both tool names and parameters, all other noise environments introduce noise to either tool names or parameters. Consequently, w
We compute the absolute difference in average LLMs performance for each type of noise added to tool names or parameters, relative to their performance in the Clean-level environment, respectively. The results depicted in Figure~\ref{fig:noise} show that tool name noise significantly affects LLMs' tool learning performance throughout the entire process. In contrast, noise in the parameters has minimal impact on the robustness of LLMs during the tool selection stage and exerts less influence on subsequent stages compared to tool name noise. Notably, LLMs exhibit greater robustness in the Union-level environment than in the Heavy (Tool) environment, underscoring the substantial impact of tool naming on model robustness.

\begin{figure}[!t]
    \centering
    \includegraphics[width=0.9\linewidth]{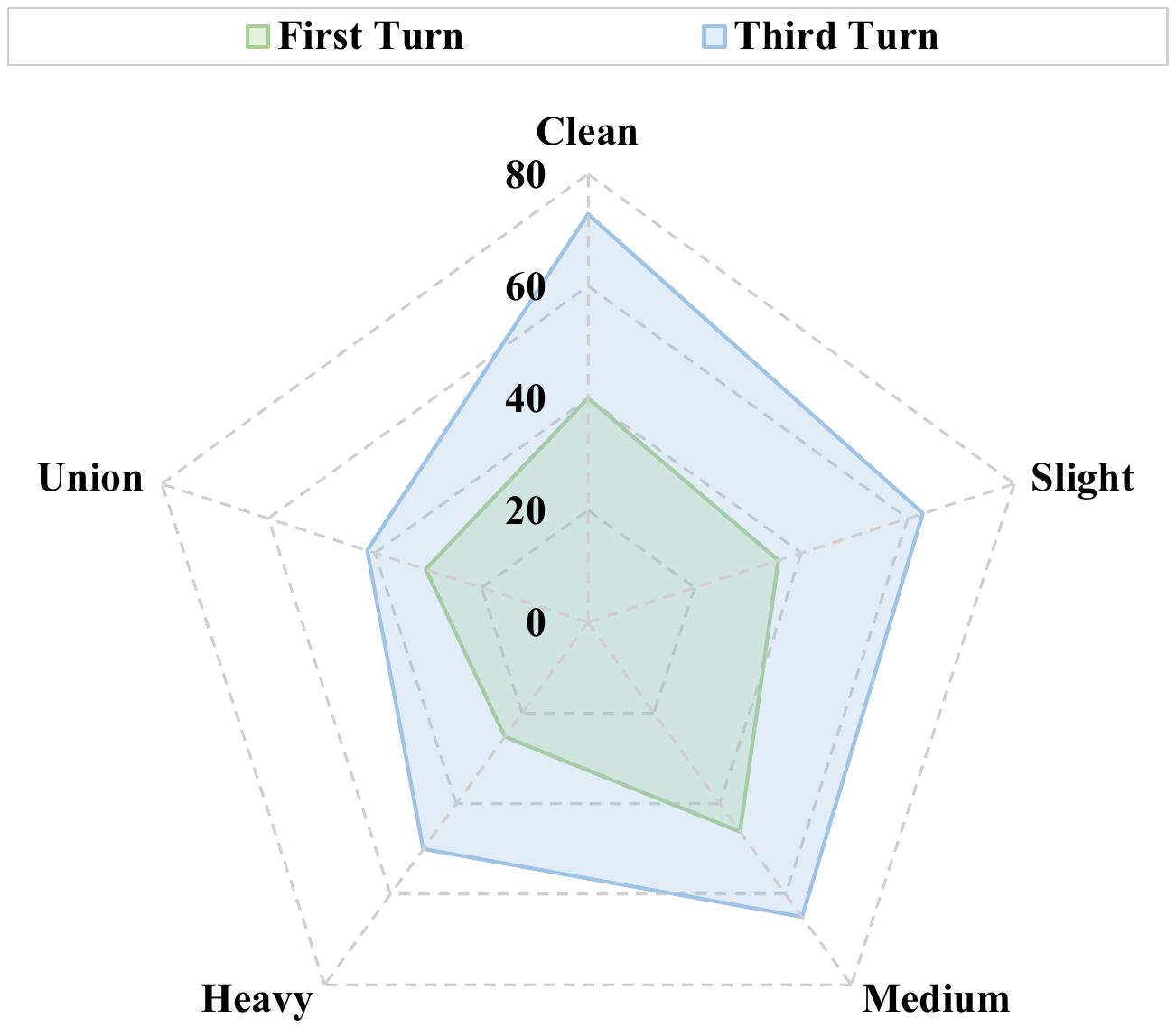}
    \caption{The performance of GPT-4 during the content filling phase in the first and third rounds of interaction.}
    \label{fig:turn3}
    % \vspace{-1mm}
\end{figure}

\textbf{Offering LLMs interactive examples enhances their tool learning performance, yet it does not bolster their robustness.}
As tool learning entails multiple turns of interaction between LLMs and external environments, we initially provide the first two turns of interactions for the test cases in each environment to evaluate LLMs' performance during the third turn of interactions.
Upon comparing GPT-4's results in the first and third turns of interactions (Figure~\ref{fig:turn3}), it becomes evident that the provision of two turns of interaction examples leads to a consistent performance boost for GPT-4, resulting in an average performance improvement of 22.91 points across various environments.
However, when examining the performance variation values, it is noteworthy that the standard deviation of its performance across environments increased from 8.14 in the first turn to 12.56 in the third turn. This observation suggests that while its performance improves, its robustness does not see a corresponding enhancement.

\begin{table}[!t]
\centering
\resizebox{\linewidth}{!}
{
\begin{tabular}{l cc} 
\toprule
\textbf{Models} & \textbf{Tool Selection} & \textbf{Parameter Identification} \\
\midrule
\textbf{GPT-3.5-turbo} & 33.72& 33.85\\
\textbf{GPT-4} & 29.17& 22.83\\
\bottomrule
\end{tabular}
}
\caption{The percentage of error caused by noise correction at different stages in GPT family of models.}
\label{tab:portion}
% \vspace{-4mm}
\end{table}

\begin{figure*}[!t]
    \centering
    \includegraphics[width=\linewidth]{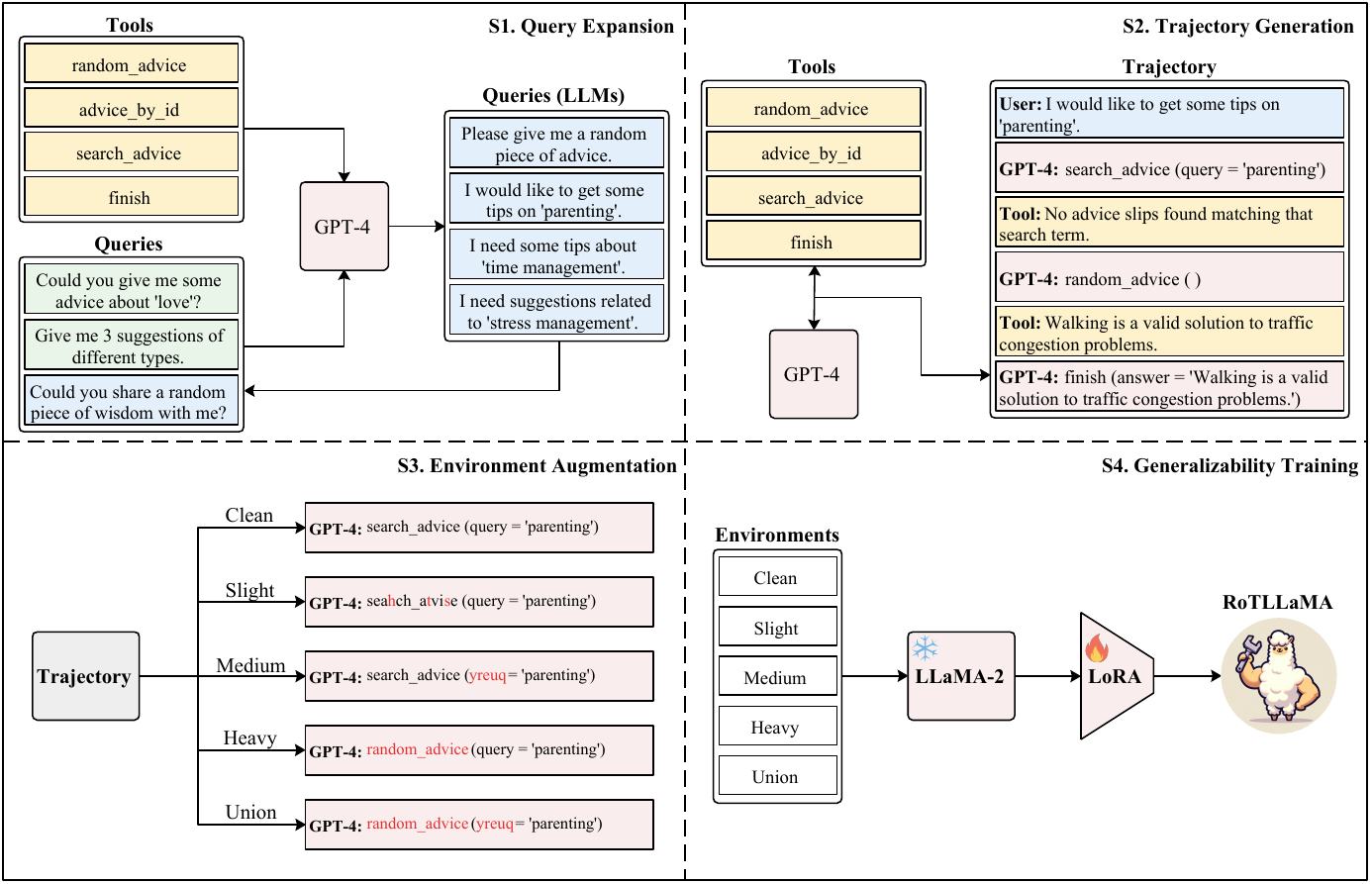}
    \caption{Illustration of RoTTuning. RoTTuning encompasses four phases, aiming at bolstering the robustness of LLMs in tool learning through increased environmental diversity.}
    \label{fig:RoTTuning}
    % \vspace{-5mm}
\end{figure*}

\subsection{Why do GPT family of models NOT perform well in Slight-level environment?}
\label{sec:gpt}

A particularly intriguing finding is that, in contrast to other LLMs, the GPT family of models exhibits a lower performance in Slight-level environment compared to Medium-level, despite the limited validity of the information provided by the latter. Our thorough investigation into the model outputs has revealed that this phenomenon can be attributed to the inherent noise correction capability of the GPT family of models. For instance, when the GPT family of models selects the tool labeled as ``predOict\_aTge,'' it automatically corrects the noise within it and generates ``predict\_age'' as the output, consequently leading to an error. \footnote{For more detailed examples, please refer to Appendix~\ref{sec:examples}.}

Table~\ref{tab:portion} illustrates the proportions of total error attributed to noise correction for the tool selection and parameter identification phases of the GPT family of models within the Slight-level environment. Notably, these proportions are exceptionally high, exceeding one-third for GPT-3.5-turbo. Consequently, addressing the challenge of mitigating capability degradation stemming from the model's inherent characteristics remains a pressing research concern.

\section{RoTTuning}

% In light of the findings, 
It is evident that enhancing the robustness of LLMs in tool learning is imperative. To tackle this issue, we introduce RoTTuning, a novel approach aimed at bolstering the robustness of LLMs through increased environmental diversity.
% Subsequent sections will provide a detailed explanation of this approach.

\subsection{Method}
% As illustrated in Figure~\ref{fig:RoTTuning}, 
RoTTuning encompasses four phases: query expansion, trajectory generation, environment augmentation, and generalizability training (Figure~\ref{fig:RoTTuning}).

\paragraph{Query Expansion}
To efficiently generate high-quality user queries on a large scale, we employ the self-instruct~\cite{self-instruct} technique, drawing from the 105 existing user queries.\footnote{The specific prompt can be found in Appendix~\ref{sec:prompt-query}.} Specifically, we instruct GPT-4 to create seven fresh user queries within the context of a subset of tools, accompanied by three existing user queries and two model-generated queries. To ensure diversity in our dataset, we scrutinize the new data for redundancy in relation to each provided example and eliminate queries with Rouge-L values surpassing 0.55. This process yields a total of 4,077 new user queries.

\paragraph{Trajectory Generation}
Upon obtaining high-quality user queries, we employ GPT-4 to produce tool learning trajectories. To ensure the accuracy of the generated trajectories, we leverage the specifically designed function call feature of GPT-4. Simultaneously, we guide GPT-4 in generating the associated thought process by incorporating a system prompt.\footnote{The specific prompt can be found in Appendix~\ref{prompt-trajectory}.} Furthermore, we specify that GPT-4's tool usage is limited to a maximum of nine turns. By considering each turn of interaction as a distinct data point, this process results in a total of 12,247 pieces of training data.

\begin{table}
\centering
\resizebox{\linewidth}{!}
{
\begin{tabular}{l| ccccc} 
\toprule
\textbf{Level} & \textbf{Clean} & \textbf{Slight} & \textbf{Medium} & \textbf{Heavy} & \textbf{Union} \\
\midrule
$\mathbf{s_{TS}}$ & 76.19 & 72.38 & 70.48 & 65.24 & 63.81 
\\
$\mathbf{s_{PI}}$ & 55.24 & 50.00 & 50.48 & 39.05 & 44.76 
\\
$\mathbf{s_{CF}}$ & 42.86 & 36.19 & 34.29 & 28.10 & 28.57 
\\
\bottomrule
\end{tabular}
}
\caption{The score in different stages (\%) of RoTLLaMA
in various Environments.}
\label{tab:RoTLLaMA}
% \vspace{-4mm}
\end{table}

\paragraph{Environment Augmentation}
% All the trajectories produced by GPT-4 are initially executed in the Clean-level environment. 
% However, 
To enhance the variety of environments,
% we aim to create trajectories that simulate noisy environments. Running GPT-4 directly in noisy environments can lead to subpar performance and compromised data quality. Therefore, 
we modify the trajectories generated in the Clean-level environment to align with the characteristics of noisy environments. This strategy ensures data quality while addressing the challenges of working in noisy settings.
To mitigate the potential drawbacks of data coupling, we introduce randomness by augmenting 3000 trajectories for each of the Slight-, Medium-, and Heavy-level environments, along with 1500 trajectories for Union-level environments. When combined with the data from the Clean-level environment, this approach yields a total of 22,747 trajectories, representing a diverse range of environmental conditions.

\paragraph{Generalizability Training}
Utilizing the diversity trajectories generated, we proceed with the fine-tuning of LLaMA-2-7B-base~\cite{LLaMA-2}
% . To enhance its capability in handling scenarios where LLMs engage in multiple turns of interaction with the external environment, we
and implement a position interpolation~\cite{interpolate} technique to extend its context length to 8096. Based on previous research indicating that fine-tuning with LoRA~\cite{LoRA} achieves superior generalization compared to full parametric fine-tuning~\cite{agenttuning}, we opt for the LoRA fine-tuning approach. We conduct 5 epochs of training to derive the ultimate model, RoTLLaMA, which exhibits robust generalization across multiple environments.

\subsection{Experimental Results}

We carry out a series of experimental analyses with RoTLLaMA on RoTBench to verify its advantages when facing various noise environments.\footnote{More experiments can be found in Appendix~\ref{sec:rottuning}.}

\paragraph{Performance}
We analyze the performance of RoTLLaMA in various environments, and the results are presented in Table~\ref{tab:RoTLLaMA}. The results reveal that RoTLLaMA's performance stability across different environments significantly surpasses that of GPT-4. Specifically, in the tool selection phase, the extreme performance difference is only 12.38, whereas GPT-4 demonstrates a much higher extreme difference of 21.90. Furthermore, in the parameter recognition and content filling phases, the extreme performance differences are 16.19 and 14.76, respectively, both of which are smaller than GPT-4's corresponding values of 20.95 and 20.95.

\begin{figure}[!t]
    \centering
    \includegraphics[width=0.9\linewidth]{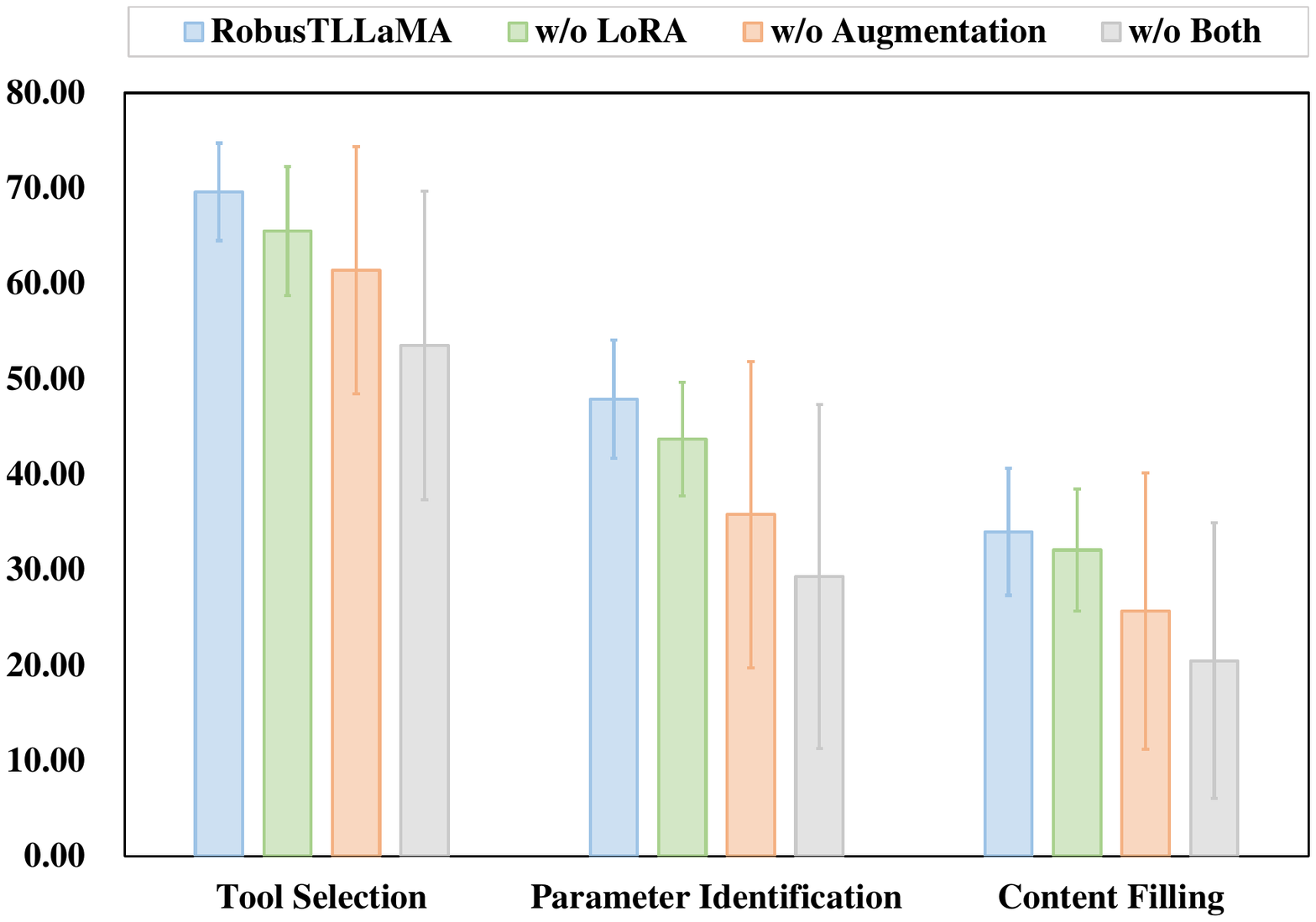}
    \caption{The means and standard deviations of our model's performance in the five environments. 
    % ``w/o'' denotes that the stage is excluded from the RoTLLaMA training process.
    }
    \label{fig:ablation}
    % \vspace{-4mm}
\end{figure}

\paragraph{Ablation Study}

To evaluate the effectiveness of various components within our approach, we conducted ablation studies on RoTLLaMA.
As shown in Figure~\ref{fig:ablation}, when substituting full-parameter fine-tuning for LoRA fine-tuning (i.e., w/o LoRA), there is a slight decrease in model performance, and standard deviations across environments remain largely unchanged. This suggests that employing LoRA enhances model performance without significantly impacting its robustness.
On the other hand, if we omit environment augmentation (i.e., w/o Augmentation), there is a notable decrease in both mean performance and a significant increase in standard deviation within each environment. This underscores the crucial role of environment augmentation in enhancing both model performance and robustness.
Furthermore, exclusively utilizing full-parameter fine-tuning on the model (i.e., w/o Both) leads to a degradation of 16.10 points in model performance.

\section{Conclusion}
In this paper, we introduce RoTBench, a multi-level benchmark for evaluating the robustness of LLMs in tool learning. RoTBench contains five environments, each characterized by varying noise levels, shedding light on the pressing need to bolster the robustness of LLMs. Furthermore, we present RoTTuning, an innovative approach that significantly improves the robustness of LLMs in tool learning by increasing the diversity of environments during the training phase.

% \newpage
% \clearpage

\section*{Limitations}
While we introduce a multi-level benchmark for evaluating the robustness of LLMs in tool learning and a training method aimed at increasing environmental diversity, our work does have some limitations.
On one hand, our primary focus is on assessing the robustness of LLMs in a single tool-use round, and we do not delve into whether LLMs are able to self-correct their behavior in response to environmental feedback. However, we analyze the performance of GPT-4 based on the interaction trajectories in the first two rounds and find that this does not enhance model robustness.
On the other hand, While tool descriptions are undoubtedly crucial for understanding tools, our analysis centers on the noise present in tool names and parameters. This choice is driven by our discovery that LLMs' comprehension of tools primarily relies on tool and parameter names rather than a nuanced understanding of the meanings conveyed in tool documentation. Within this framework, evaluating LLMs through RoTBench can effectively measure their tolerance to noise in these additional details, thus propelling research endeavors aimed at improving LLMs' tool learning capabilities.

\section*{Acknowledgements}
The authors wish to thank the anonymous reviewers for their helpful comments. This work was partially funded by National Natural Science Foundation of China (No. 62476061,62206057,62076069), Shanghai Rising-Star Program (23QA1400200), Natural Science Foundation of Shanghai (23ZR1403500), Program of Shanghai Academic Research Leader under grant 22XD1401100.

\bibliography{anthology,custom}

\clearpage

\appendix
\section{Details of LLMs}
\label{sec:models}

To evaluate the robustness of widely-used LLMs with tool-use capabilities, we opt for testing four open-source models and two closed-source models.

\subsection{Open-Source LLMs}
Among open-source LLMs, we have chosen four models that have undergone dedicated training for tool learning.

\paragraph{ToolLLaMA-2-7B-v1}
ToolLLaMA-2-7B-v1, developed by Tsinghua University, is a tool-oriented LLM that harnesses the power of 126,000 data samples, including more than 16,000 APIs, through supervised fine-tuning on LLaMA-2-7B-base. This enables ToolLLaMA-2-7B-v1 to effectively utilize various tools to meet diverse user requirements.

\paragraph{ToolLLaMA-2-7B-v2}
ToolLLaMA-2-7B-v2 has undergone fine-tuning from LLaMA-2-7B-base, by assimilating an expansive dataset comprising over 120,000 solution paths and annotated chains of thought. To the best of our knowledge, this model stands as the most extensively trained tool-oriented LLM, utilizing the largest dataset and the broadest spectrum of tools among all available options.

\paragraph{NexusRaven-13B-v1}
NexusRaven-13B-v1 is a tool-oriented model that underwent fine-tuning based on CodeLLaMA-13B. Distinguishing itself from prior models, NexusRaven-13B-v1 employs code nesting to invoke tools, generating the entire inference path simultaneously instead of following a step-by-step approach.

\paragraph{NexusRaven-13B-v2}
NexusRaven-13B-v2 enhances the performance of NexusRaven-13B-v1 by generating single, nested, and parallel function calls in various complex scenarios. Additionally, NexusRaven-13B-v2 can generate inference paths for the function calls it creates, thereby improving overall generalization.

\subsection{Closed-Source LLMs}
Among closed-source LLMs, we have opted for two of the most representative models from the GPT family.

\paragraph{GPT-3.5-turbo}
GPT-3.5-turbo stands out as the most potent and cost-efficient model within the GPT-3.5 series. Tailored for conversations, it excels in comprehending and generating natural language. Furthermore, it exhibits strong tool invocation capabilities.

\paragraph{GPT-4}
GPT-4 represents OpenAI's most robust LLM, surpassing its predecessor in delivering safer and more beneficial responses. Additionally, GPT-4 offers formal support for multimodal inputs and has an expanded capability to address a broader spectrum of social requirements.

\section{Experimental Setup}
\paragraph{Inference}
In accordance with~\citet{ToolEyes}, we adopt the ReAct~\cite{React} format for inference, employing a consistent prompt template for both the ToolLLaMA-2-7B family of models and the GPT family of models. However, as NexusRaven-13B fmaily of models utilize nested functions for output, we adhere to the guidelines outlined on their official website, which necessitate the use of a distinct set of template.\footnote{The specific prompt can be found in Appendix~\ref{sec:prompt}.}
Meanwhile, to evaluate human performance across environments with different noise levels, we enlist three university students. Each student receives identical tool documentation and task descriptions. Independently, they completes the questions and the average score derived from their responses served as the human performance benchmark.

\paragraph{Evaluation}
We score the performance of LLMs and Human using the evaluation methods defined in Section~\ref{sec:eval}. In this system, each data point is scored as 0 or 1 at each stage. This is because, in the context of tool learning, tool calls either succeed or fail, and even small errors can cause the entire process to fail. 
In particular, In the tool selection phase, an error in tool selection can lead to overall failure, independent of parameter accuracy. In the parameter identification phase, missing necessary parameters or wrong parameter selection can lead to failure. In the content filling phase, incorrect content input can lead to undesirable tool execution results.

\section{Results in Different Scenarios}
\label{sec:scenarios}
We show the performance of each model in different scenarios and document the results from Table~\ref{tab:TG} to Table~\ref{tab:FT}.
From the results, we have the following observations.

\paragraph{The variance in average performance of LLMs across various study scenarios can be attributed to the relevance of specific features of available tools to each scenario.}
For instance, in both application operations and personal life scenarios, LLMs may err due to the strict sequential order in which tools are called (e.g., obtaining parameter values for ``list\_properties'' necessitates prior execution of ``search\_locations'').

\paragraph{It's notable that the model's perception of environmental complexity may diverge from human intentions.}
For instance, in information retrieval scenarios, LLMs exhibit inferior average performance in the slight-level environment compared to the medium-level and heavy-level environments, primarily due to limitations in noise correction capabilities (Section~\ref{sec:gpt}).

\paragraph{Regarding the model itself, variations in training methods and data can lead to unexpected performances in certain scenarios.}
For instance, ToolLLaMA-7B-v1 demonstrates a performance discrepancy between the clean-level and union-level environments in the application manipulation scenario, scoring 20 and 40, respectively. This disparity arises from its ability to perform better when only two tools are available alongside ``ask\_to\_user'' and ``finish,'' whereas GPT4 consistently prompts for API keys even when unnecessary.

\begin{table*}[!t]
\centering
\resizebox{\linewidth}{!}
{
\begin{tabular}{l | c c |c c |c c}
\toprule
\multirow{3}{0.1\linewidth}{\textbf{Models}} & \multicolumn{4}{c|}{\textbf{Open-Source LLMs}} & \multicolumn{2}{c}{\textbf{Closed-Source LLMs}}\\ \cmidrule(lr){2-5} \cmidrule(lr){6-7}

& \multirow{2}{0.13\linewidth}{\centering \textbf{ToolLLaMA-2-7B-v1}} & \multirow{2}{0.13\linewidth}{\centering \textbf{ToolLLaMA-2-7B-v2}} & \multirow{2}{0.13\linewidth}{\centering \textbf{NexusRaven-13B-v1}} & \multirow{2}{0.13\linewidth}{\centering \textbf{NexusRaven-13B-v2}} & \multirow{2}{0.13\linewidth}{\centering \textbf{GPT-3.5-turbo}} & \multirow{2}{0.13\linewidth}{\centering \textbf{GPT-4}} \\
&&&&&&\\
\midrule
\multicolumn{7}{c}{\textit{Tool Selection}} \\
\midrule
\rowcolor{gray!10} \textbf{Clean} & 60.00& 73.33& 20.00& 53.33& \textbf{86.67}& \textbf{86.67}\\
\textbf{Slight} & 46.67& 60.00& 30.00& 56.67& 73.33& \textbf{83.33}\\
\textbf{Medium} & 36.67& 50.00& 30.00& 70.00& 73.33& \textbf{90.00}\\
\textbf{Heavy} & 36.67& 43.33& 20.00& 40.00& 53.33& \textbf{70.00}\\
\textbf{Union} & 40.00& 26.67& 26.67& 46.67& \textbf{60.00}& 46.67\\
\midrule
\multicolumn{7}{c}{\textit{Parameter Identification}} \\
\midrule
\rowcolor{gray!10} \textbf{Clean} & 60.00& 60.00& 6.67& 40.00& 60.00& \textbf{73.33}\\
\textbf{Slight} & 40.00& 46.67& 13.33& 40.00& 36.67& \textbf{53.33}\\
\textbf{Medium} & 33.33& 40.00& 10.00& 50.00& 40.00& \textbf{63.33}\\
\textbf{Heavy} & 36.67& 30.00& 6.67& 13.33& 23.33& \textbf{40.00}\\
\textbf{Union} & \textbf{40.00}& 13.33& 13.33& \textbf{40.00}& 26.67& 33.33\\
\midrule
\multicolumn{7}{c}{\textit{Content Filling}} \\
\midrule
\rowcolor{gray!10} \textbf{Clean} & 26.67& 26.67& 6.67& 33.33& 60.00& \textbf{73.33}
\\
\textbf{Slight} & 16.67& 13.33& 10.00& 33.33& 36.67& \textbf{53.33}
\\
\textbf{Medium} & 13.33& 10.00& 6.67& 36.67& 40.00& \textbf{63.33}
\\
\textbf{Heavy} & 16.67& 13.33& 3.33& 13.33& 20.00& \textbf{36.67}
\\
\textbf{Union} & 20.00& 0.00& 6.67& \textbf{33.33}& 26.67& \textbf{33.33}\\
\bottomrule
\end{tabular}
}
\caption{Performance of various LLMs in the text generation scenario, with the best performance in each environment highlighted in \textbf{bold}.}
\label{tab:TG}

\end{table*}

\begin{table*}[!t]
\centering
\resizebox{\linewidth}{!}
{
\begin{tabular}{l  |c c |c c |c c }
\toprule
\multirow{3}{0.1\linewidth}{\textbf{Models}} & \multicolumn{4}{c|}{\textbf{Open-Source LLMs}} & \multicolumn{2}{c}{\textbf{Closed-Source LLMs}} \\ \cmidrule(lr){2-5} \cmidrule(lr){6-7}

& \multirow{2}{0.13\linewidth}{\centering \textbf{ToolLLaMA-2-7B-v1}} & \multirow{2}{0.13\linewidth}{\centering \textbf{ToolLLaMA-2-7B-v2}} & \multirow{2}{0.13\linewidth}{\centering \textbf{NexusRaven-13B-v1}} & \multirow{2}{0.13\linewidth}{\centering \textbf{NexusRaven-13B-v2}} & \multirow{2}{0.13\linewidth}{\centering \textbf{GPT-3.5-turbo}} & \multirow{2}{0.13\linewidth}{\centering \textbf{GPT-4}} \\
&&&&&&\\
\midrule
\multicolumn{7}{c}{\textit{Tool Selection}}\\
\midrule
\rowcolor{gray!10} \textbf{Clean} & 80.00& 80.00& 80.00& 80.00& \textbf{86.67}& \textbf{86.67}
\\
\textbf{Slight} & 63.33& 80.00& 70.00& \textbf{83.33}& 63.33& 73.33
\\
\textbf{Medium} & 60.00& 73.33& 66.67& 80.00& 83.33& \textbf{93.33}
\\
\textbf{Heavy} & 46.67& 56.67& 50.00& \textbf{60.00}& 56.67& 56.67
\\
\textbf{Union} & 40.00& 53.33& 46.67& 60.00& 60.00& \textbf{86.67}\\
\midrule
\multicolumn{7}{c}{\textit{Parameter Identification}}\\
\midrule
\rowcolor{gray!10} \textbf{Clean} & 60.00& 40.00& 26.67& 33.33& 40.00& \textbf{66.67}
\\
\textbf{Slight} & 50.00& 43.33& 26.67& 36.67& 26.67& \textbf{60.00}
\\
\textbf{Medium} & 50.00& 46.67& 16.67& 30.00& 40.00& \textbf{66.67}
\\
\textbf{Heavy} & 33.33& \textbf{40.00}& 10.00& 26.67& 13.33& 26.67
\\
\textbf{Union} & 20.00& 46.67& 6.67& 20.00& 13.33& \textbf{60.00}\\
\midrule
\multicolumn{7}{c}{\textit{Content Filling}}\\
\midrule
\rowcolor{gray!10} \textbf{Clean} & 46.67& 33.33& 0.00& 20.00& 26.67& \textbf{53.33}
\\
\textbf{Slight} & 33.33& 40.00& 0.00& 23.33& 16.67& \textbf{53.33}
\\
\textbf{Medium} & 30.00& 40.00& 0.00& 16.67& 30.00& \textbf{56.67}
\\
\textbf{Heavy} & 13.33& 20.00& 0.00& \textbf{23.33}& 10.00& 20.00
\\
\textbf{Union} & 13.33& 40.00& 0.00& 13.33& 6.67& \textbf{46.67}\\
\bottomrule
\end{tabular}
}
\caption{Performance of various LLMs in the data understanding scenario, with the best performance in each environment highlighted in \textbf{bold}.}
\label{tab:DU}

\end{table*}

\begin{table*}[!t]
\centering
\resizebox{\linewidth}{!}
{
\begin{tabular}{l  |c c |c c |c c }
\toprule
\multirow{3}{0.1\linewidth}{\textbf{Models}} & \multicolumn{4}{c|}{\textbf{Open-Source LLMs}} & \multicolumn{2}{c}{\textbf{Closed-Source LLMs}} \\ \cmidrule(lr){2-5} \cmidrule(lr){6-7}

& \multirow{2}{0.13\linewidth}{\centering \textbf{ToolLLaMA-2-7B-v1}} & \multirow{2}{0.13\linewidth}{\centering \textbf{ToolLLaMA-2-7B-v2}} & \multirow{2}{0.13\linewidth}{\centering \textbf{NexusRaven-13B-v1}} & \multirow{2}{0.13\linewidth}{\centering \textbf{NexusRaven-13B-v2}} & \multirow{2}{0.13\linewidth}{\centering \textbf{GPT-3.5-turbo}} & \multirow{2}{0.13\linewidth}{\centering \textbf{GPT-4}} \\
&&&&&&\\
\midrule
\multicolumn{7}{c}{\textit{Tool Selection}}\\
\midrule
\rowcolor{gray!10} \textbf{Clean} & 66.67& 60.00& 40.00& 86.67& 73.33& \textbf{93.33}
\\
\textbf{Slight} & 60.00& 50.00& 36.67& \textbf{80.00}& 60.00& \textbf{80.00}
\\
\textbf{Medium} & 63.33& 46.67& 43.33& 76.67& 73.33& \textbf{90.00}
\\
\textbf{Heavy} & 46.67& 36.67& 36.67& \textbf{73.33}& 46.67& 56.67
\\
\textbf{Union} & 53.33& 46.67& 26.67& 66.67& 60.00& \textbf{73.33}\\
\midrule
\multicolumn{7}{c}{\textit{Parameter Identification}}\\
\midrule
\rowcolor{gray!10} \textbf{Clean} & 60.00& 46.67& 6.67& \textbf{73.33}& 53.33& 53.33
\\
\textbf{Slight} & 53.33& 43.33& 6.67& \textbf{66.67}& 36.67& 40.00
\\
\textbf{Medium} & 46.67& 40.00& 10.00& \textbf{60.00}& 53.33& 53.33
\\
\textbf{Heavy} & 30.00& 30.00& 6.67& \textbf{43.33}& 16.67& 23.33
\\
\textbf{Union} & \textbf{40.00}& 33.33& 6.67& \textbf{40.00}& 33.33& \textbf{40.00}\\
\midrule
\multicolumn{7}{c}{\textit{Content Filling}}\\
\midrule
\rowcolor{gray!10} \textbf{Clean} & \textbf{33.33}& 20.00& 0.00& \textbf{33.33}& 20.00& \textbf{33.33}
\\
\textbf{Slight} & \textbf{30.00}& 20.00& 0.00& \textbf{30.00}& 20.00& \textbf{30.00}
\\
\textbf{Medium} & 16.67& 10.00& 0.00& 26.67& 30.00& \textbf{40.00}
\\
\textbf{Heavy} & 6.67& 20.00& 0.00& \textbf{26.67}& 10.00& 20.00
\\
\textbf{Union} & 13.33& 13.33& 0.00& 6.67& 26.67& \textbf{40.00}\\
\bottomrule
\end{tabular}
}
\caption{Performance of various LLMs in the real-time search scenario, with the best performance in each environment highlighted in \textbf{bold}.}
\label{tab:RS}

\end{table*}

\begin{table*}[!t]
\centering
\resizebox{\linewidth}{!}
{
\begin{tabular}{l  |c c |c c |c c }
\toprule
\multirow{3}{0.1\linewidth}{\textbf{Models}} & \multicolumn{4}{c|}{\textbf{Open-Source LLMs}} & \multicolumn{2}{c}{\textbf{Closed-Source LLMs}} \\ \cmidrule(lr){2-5} \cmidrule(lr){6-7}

& \multirow{2}{0.13\linewidth}{\centering \textbf{ToolLLaMA-2-7B-v1}} & \multirow{2}{0.13\linewidth}{\centering \textbf{ToolLLaMA-2-7B-v2}} & \multirow{2}{0.13\linewidth}{\centering \textbf{NexusRaven-13B-v1}} & \multirow{2}{0.13\linewidth}{\centering \textbf{NexusRaven-13B-v2}} & \multirow{2}{0.13\linewidth}{\centering \textbf{GPT-3.5-turbo}} & \multirow{2}{0.13\linewidth}{\centering \textbf{GPT-4}} \\
&&&&&&\\
\midrule
\multicolumn{7}{c}{\textit{Tool Selection}}\\
\midrule
\rowcolor{gray!10} \textbf{Clean} & \textbf{86.67}& 73.33& 73.33& 66.67& 80.00& 73.33
\\
\textbf{Slight} & \textbf{80.00}& \textbf{80.00}& 73.33& 70.00& 66.67& 73.33
\\
\textbf{Medium} & 83.33& 80.00& 73.33& 66.67& 80.00& \textbf{86.67}
\\
\textbf{Heavy} & 60.00& 50.00& \textbf{70.00}& 66.67& \textbf{70.00}& 63.33
\\
\textbf{Union} & \textbf{80.00}& 53.33& 73.33& 66.67& 66.67& 53.33\\
\midrule
\multicolumn{7}{c}{\textit{Parameter Identification}}\\
\midrule
\rowcolor{gray!10} \textbf{Clean} & 40.00& 40.00& 6.67& \textbf{60.00}& 53.33& 46.67
\\
\textbf{Slight} & 56.67& 46.67& 10.00& \textbf{60.00}& 36.67& 46.67
\\
\textbf{Medium} & 53.33& 46.67& 6.67& 53.33& \textbf{56.67}& 46.67
\\
\textbf{Heavy} & 36.67& 20.00& 13.33& \textbf{50.00}& 40.00& 43.33
\\
\textbf{Union} & \textbf{73.33}& 40.00& 13.33& 53.33& 40.00& 33.33\\
\midrule
\multicolumn{7}{c}{\textit{Content Filling}}\\
\midrule
\rowcolor{gray!10} \textbf{Clean} & \textbf{20.00}& 13.33& 0.00& \textbf{20.00}& \textbf{20.00}&\textbf{20.00}
\\
\textbf{Slight} & \textbf{33.33}& 20.00& 0.00& 20.00& 16.67& 13.33
\\
\textbf{Medium} & \textbf{40.00}& 26.67& 0.00& 16.67& 26.67& 23.33
\\
\textbf{Heavy} & 20.00& 6.67& 0.00& \textbf{26.67}& 16.67& 13.33
\\
\textbf{Union} & \textbf{40.00}& 26.67& 0.00& 13.33& 20.00& 6.67\\
\bottomrule
\end{tabular}
}
\caption{Performance of various LLMs in the application manipulation scenatio, with the best performance in each environment highlighted in \textbf{bold}.}
\label{tab:AM}

\end{table*}

\begin{table*}[!t]
\centering
\resizebox{\linewidth}{!}
{
\begin{tabular}{l  |c c |c c |c c }
\toprule
\multirow{3}{0.1\linewidth}{\textbf{Models}} & \multicolumn{4}{c|}{\textbf{Open-Source LLMs}} & \multicolumn{2}{c}{\textbf{Closed-Source LLMs}} \\ \cmidrule(lr){2-5} \cmidrule(lr){6-7}

& \multirow{2}{0.13\linewidth}{\centering \textbf{ToolLLaMA-2-7B-v1}} & \multirow{2}{0.13\linewidth}{\centering \textbf{ToolLLaMA-2-7B-v2}} & \multirow{2}{0.13\linewidth}{\centering \textbf{NexusRaven-13B-v1}} & \multirow{2}{0.13\linewidth}{\centering \textbf{NexusRaven-13B-v2}} & \multirow{2}{0.13\linewidth}{\centering \textbf{GPT-3.5-turbo}} & \multirow{2}{0.13\linewidth}{\centering \textbf{GPT-4}} \\
&&&&&&\\
\midrule
\multicolumn{7}{c}{\textit{Tool Selection}}\\
\midrule
\rowcolor{gray!10} \textbf{Clean} & 53.33& 60.00& 40.00& 66.67& \textbf{73.33}& 66.67
\\
\textbf{Slight} & 46.67& 63.33& 43.33& \textbf{73.33}& 50.00& 70.00
\\
\textbf{Medium} & 50.00& 53.33& 50.00& 63.33& 60.00& \textbf{73.33}
\\
\textbf{Heavy} & 23.33& 40.00& 43.33& \textbf{50.00}& \textbf{50.00}& \textbf{50.00}
\\
\textbf{Union} & 40.00& \textbf{53.33}& \textbf{53.33}& 46.67& 40.00& 46.67\\
\midrule
\multicolumn{7}{c}{\textit{Parameter Identification}}\\
\midrule
\rowcolor{gray!10} \textbf{Clean} & 26.67& 40.00& 13.33& \textbf{53.33}& 26.67& 40.00
\\
\textbf{Slight} & 30.00& 26.67& 13.33& \textbf{53.33}& 10.00& 26.67
\\
\textbf{Medium} & 26.67& 26.67& 13.33& 36.67& \textbf{40.00}& \textbf{40.00}
\\
\textbf{Heavy} & 6.67& 16.67& 3.33& \textbf{30.00}& 16.67& 26.67
\\
\textbf{Union} & 26.67& 20.00& 6.67& 26.67& 26.67& \textbf{40.00}\\
\midrule
\multicolumn{7}{c}{\textit{Content Filling}}\\
\midrule
\rowcolor{gray!10} \textbf{Clean} & 20.00& 26.67& 0.00& \textbf{40.00}& 13.33& 33.33
\\
\textbf{Slight} & 16.67& 20.00& 0.00& \textbf{43.33}& 10.00& 23.33
\\
\textbf{Medium} & 13.33& 23.33& 0.00& 33.33& 30.00& \textbf{40.00}
\\
\textbf{Heavy} & 6.67& 10.00& 0.00& \textbf{26.67}& 10.00& \textbf{26.67}
\\
\textbf{Union} & 6.67& 20.00& 0.00& \textbf{26.67}& 6.67& \textbf{26.67}\\
\bottomrule
\end{tabular}
}
\caption{Performance of various LLMs in the personal life scenario, with the best performance in each environment highlighted in \textbf{bold}.}
\label{tab:PL}

\end{table*}

\begin{table*}[!t]
\centering
\resizebox{\linewidth}{!}
{
\begin{tabular}{l  |c c |c c |c c }
\toprule
\multirow{3}{0.1\linewidth}{\textbf{Models}} & \multicolumn{4}{c|}{\textbf{Open-Source LLMs}} & \multicolumn{2}{c}{\textbf{Closed-Source LLMs}} \\ \cmidrule(lr){2-5} \cmidrule(lr){6-7}

& \multirow{2}{0.13\linewidth}{\centering \textbf{ToolLLaMA-2-7B-v1}} & \multirow{2}{0.13\linewidth}{\centering \textbf{ToolLLaMA-2-7B-v2}} & \multirow{2}{0.13\linewidth}{\centering \textbf{NexusRaven-13B-v1}} & \multirow{2}{0.13\linewidth}{\centering \textbf{NexusRaven-13B-v2}} & \multirow{2}{0.13\linewidth}{\centering \textbf{GPT-3.5-turbo}} & \multirow{2}{0.13\linewidth}{\centering \textbf{GPT-4}} \\
&&&&&&\\
\midrule
\multicolumn{7}{c}{\textit{Tool Selection}}\\
\midrule
\rowcolor{gray!10} \textbf{Clean} & 60.00& \textbf{80.00}& 73.33& 73.33& 46.67& 73.33
\\
\textbf{Slight} & 50.00& 63.33& 66.67& \textbf{83.33}& 43.33& 73.33
\\
\textbf{Medium} & 43.33& 56.67& 63.33& \textbf{76.67}& 53.33& 73.33
\\
\textbf{Heavy} & 50.00& 53.33& 53.33& \textbf{80.00}& 53.33& 56.67
\\
\textbf{Union} & 26.67& 33.33& 46.67& \textbf{53.33}& 40.00& 40.00\\
\midrule
\multicolumn{7}{c}{\textit{Parameter Identification}}\\
\midrule
\rowcolor{gray!10} \textbf{Clean} & 26.67& 33.33& 26.67& \textbf{53.33}& 40.00& 40.00
\\
\textbf{Slight} & 16.67& 20.00& 23.33& \textbf{60.00}& 30.00& 36.67
\\
\textbf{Medium} & 16.67& 16.67& 30.00& \textbf{60.00}& 43.33& 50.00
\\
\textbf{Heavy} & 23.33& 26.67& 16.67& \textbf{56.67}& 33.33& 36.67
\\
\textbf{Union} & 20.00& 13.33& 20.00& \textbf{40.00}& \textbf{40.00}& \textbf{40.00}\\
\midrule
\multicolumn{7}{c}{\textit{Content Filling}}\\
\midrule
\rowcolor{gray!10} \textbf{Clean} & 20.00& 26.67& 0.00& \textbf{46.67}& 26.67& 33.33
\\
\textbf{Slight} & 13.33& 16.67& 6.67& \textbf{56.67}& 23.33& 30.00
\\
\textbf{Medium} & 16.67& 13.33& 3.33& \textbf{53.33}& 33.33& 46.67
\\
\textbf{Heavy} & 23.33& 16.67& 3.33& \textbf{53.33}& 26.67& 30.00
\\
\textbf{Union} & 13.33& 6.67& 0.00& \textbf{33.33}& \textbf{33.33}& \textbf{33.33}\\
\bottomrule
\end{tabular}
}
\caption{Performance of various LLMs in the information retrieval scenario, with the best performance in each environment highlighted in \textbf{bold}.}
\label{tab:IR}

\end{table*}

\begin{table*}[!t]
\centering
\resizebox{\linewidth}{!}
{
\begin{tabular}{l  |c c |c c |c c }
\toprule
\multirow{3}{0.1\linewidth}{\textbf{Models}} & \multicolumn{4}{c|}{\textbf{Open-Source LLMs}} & \multicolumn{2}{c}{\textbf{Closed-Source LLMs}} \\ \cmidrule(lr){2-5} \cmidrule(lr){6-7}

& \multirow{2}{0.13\linewidth}{\centering \textbf{ToolLLaMA-2-7B-v1}} & \multirow{2}{0.13\linewidth}{\centering \textbf{ToolLLaMA-2-7B-v2}} & \multirow{2}{0.13\linewidth}{\centering \textbf{NexusRaven-13B-v1}} & \multirow{2}{0.13\linewidth}{\centering \textbf{NexusRaven-13B-v2}} & \multirow{2}{0.13\linewidth}{\centering \textbf{GPT-3.5-turbo}} & \multirow{2}{0.13\linewidth}{\centering \textbf{GPT-4}} \\
&&&&&&\\
\midrule
\multicolumn{7}{c}{\textit{Tool Selection}}\\
\midrule
\rowcolor{gray!10} \textbf{Clean} & 46.67& 53.33& 53.33& \textbf{73.33}& 66.67& 66.67
\\
\textbf{Slight} & 43.33& 50.00& 43.33& \textbf{73.33}& 43.33& \textbf{73.33}
\\
\textbf{Medium} & 46.67& 43.33& 40.00& 66.67& 50.00& \textbf{70.00}
\\
\textbf{Heavy} & 26.67& 36.67& 36.67& \textbf{53.33}& 50.00& \textbf{53.33}
\\
\textbf{Union} & 20.00& 26.67& 26.67& \textbf{46.67}& 33.33& \textbf{46.67}\\
\midrule
\multicolumn{7}{c}{\textit{Parameter Identification}}\\
\midrule
\rowcolor{gray!10} \textbf{Clean} & 33.33& 33.33& 20.00& \textbf{66.67}& 60.00& 40.00
\\
\textbf{Slight} & 26.67& 40.00& 23.33& \textbf{66.67}& 20.00& 46.67
\\
\textbf{Medium} & 26.67& 23.33& 16.67& \textbf{56.67}& 36.67& 50.00
\\
\textbf{Heavy} & 16.67& 16.67& 13.33& \textbf{33.33}& 26.67& 23.33
\\
\textbf{Union} & 13.33& 13.33& 13.33& \textbf{33.33}& 13.33& 26.67\\
\midrule
\multicolumn{7}{c}{\textit{Content Filling}}\\
\midrule
\rowcolor{gray!10} \textbf{Clean} & 33.33& 33.33& 6.67& \textbf{60.00}& 46.67& 33.33
\\
\textbf{Slight} & 26.67& 36.67& 6.67& \textbf{60.00}& 16.67& 46.67
\\
\textbf{Medium} & 26.67& 23.33& 3.33& \textbf{46.67}& 23.33& \textbf{46.67}
\\
\textbf{Heavy} & 13.33& 16.67& 0.00& \textbf{33.33}& 20.00& 23.33
\\
\textbf{Union} & 6.67& 6.67& 6.67& \textbf{26.67}& 6.67& \textbf{26.67}\\
\bottomrule
\end{tabular}
}
\caption{Performance of various LLMs in the financial transactions scenario, with the best performance in each environment highlighted in \textbf{bold}.}
\label{tab:FT}

\end{table*}

\section{Examples for Noise Correction}
\label{sec:examples}
In Table~\ref{tab:noise-correct}, we present instances of noise correction observed during the tool selection and parameter identification phases of the GPT family of models.

\begin{table*}[!t]
    \centering
    \resizebox{\linewidth}{!}
    {
        \begin{tabular}{m{0.1\linewidth}|m{0.14\linewidth}|m{0.22\linewidth}|m{0.28\linewidth}|m{0.26\linewidth}}
        \toprule
         \textbf{Models} & \multicolumn{1}{l|}{\textbf{Stage}} & \multicolumn{1}{c|}{\textbf{Query}} & \multicolumn{1}{c|}{\textbf{Noisy Part}} & \multicolumn{1}{c}{\textbf{Model Output}}  \\ \midrule
         \textbf{GPT-3.5-turbo}& \textbf{Tool Selection} & I have a list of names: Maria, Juan, and Carlos. Can you predict their ages? & 
         Tool: \textcolor{red}{predOict\_aTge}
         
         Description: Predicts the ages of one or more people given their names.

         Parameters: ...
         &
Tool: \textcolor{red}{predict\_age}
\\
\midrule
\textbf{GPT-3.5-turbo}& \textbf{Parameter Identification} &   I want to know what will be the output if we run these commands sequentially in bash: `cd /home/user/documents', `ls -a.' &
Tool: execute\_bash\_code

Description: ...

Parameters: \textcolor{red}{Nommands} (Required)

Param Description: The command string to be executed.
& 
Parameters: \textcolor{red}{commands} \\
\midrule
\textbf{GPT-4} & \textbf{Tool Selection} & Is there any social event available which requires high accessibility and is free of cost?
& Tool: \textcolor{red}{get\_activty\_by\_ye}

Description: Find a random activity with a given type.

Parameters: ...
& Tool: \textcolor{red}{get\_activity\_by\_type}\\
\midrule
\textbf{GPT-4} &\textbf{Parameter Identification} & Get me quotes for symbols AAPL, MSFT, and GOOGL from US.& Tool: get\_quotes

Description: ...

Parameters: \textcolor{red}{ymbols} (Required)

Param Description: The value of symbol field returned in auto-complete endpoint. Separated by comma for multiple entities.& Parameters: \textcolor{red}{symbols} \\
\bottomrule
    \end{tabular}
    }
    \caption{Examples for noise correction of GPT family of models.}
    \label{tab:noise-correct}
\end{table*}

\section{Further Studies about RoTTuning}
\label{sec:rottuning}

We conduct additional comparative analysis to further validate the effectiveness of RoTTuning in improving the stability of LLMs in noisy environments.

\paragraph{Robust Generalization of RoTTuning}
To validate the robust generalization of RoTTuning across different environments, we apply a single environment augmentation and compare the results to those without augmentation. As shown in Table ~\ref{tab:RoTTuning}, even when training RoTTuning with data from only one environment, it achieves superior performance in other environments, demonstrating its strong generalization capability.

\begin{table*}[!t]
\centering
\resizebox{\linewidth}{!}
{
\begin{tabular}{l| ccccc} 
\toprule
 \textbf{Approaches} & \textbf{w/o Augmentation} & \textbf{w/ Aug.Slight} & \textbf{w/ Aug.Medium} & \textbf{w/ Aug.Heavy} & \textbf{w/ Aug.Union} \\
\midrule
\multicolumn{6}{c}{\textit{Tool Selection}} \\
\midrule
\rowcolor{gray!10} \textbf{Clean} & 74.29& 70.48& 72.38& \textbf{75.24}& 71.43\\
\textbf{Slight} & 65.24& \textbf{71.90}& 62.38& 69.05& 64.29\\
\textbf{Medium} & 61.90& 68.57& 65.71& \textbf{70.95}& 66.67\\
\textbf{Heavy} & 50.48& 51.90& 49.52& \textbf{60.48}& 55.24\\
\textbf{Union} & 40.00& 53.33& 51.43& 53.33& \textbf{55.24}\\
\midrule
\multicolumn{6}{c}{\textit{Parameter Identification}} \\
\midrule
\rowcolor{gray!10} \textbf{Clean} & \textbf{60.95}& 57.14& 59.05& 59.05& \textbf{60.95}\\
\textbf{Slight} & 47.14& \textbf{53.81}& 46.19& 48.10& 46.19\\
\textbf{Medium} & 42.86& 51.90& 48.57& 48.57& \textbf{52.38}\\
\textbf{Heavy} & 14.29& 18.10& 15.24& \textbf{33.81}& 26.67\\
\textbf{Union} & 21.90& 32.38& 28.57& 31.43& \textbf{36.19}\\
\midrule
\multicolumn{6}{c}{\textit{Content Filling}} \\
\midrule
\rowcolor{gray!10} \textbf{Clean} & 45.71& 43.81& \textbf{48.57}& 44.76& 42.86\\
\textbf{Slight} & 31.90& \textbf{40.00}& 31.90& 35.24& 30.95\\
\textbf{Medium} & 30.48& 38.10& 36.67& 36.67& \textbf{38.57}\\
\textbf{Heavy} & 10.48& 12.86& 10.48& \textbf{24.7}6& 19.05\\
\textbf{Union} & 12.38& 19.05& 17.14& 21.90& \textbf{27.62}\\
\bottomrule
\end{tabular}
}
\caption{Performance of the LLMs trained by data augmented from single environment, compared with the model trained using LoRA without augmentation. The best performance in each environment is highlighted in \textbf{bold}.}
\label{tab:RoTTuning}
\end{table*}

\paragraph{The Number of Tool Hallucinations} 
We compare the number of tool hallucinations for each LLM in all environments and find that our model has significantly fewer hallucinations compared to the GPT family of models (Table~\ref{tab:hallucinations}). This demonstrates the effectiveness of our method in mitigating interference from various sources of noise while accurately acquiring environmental information. It's worth noting that the NexusRaven family of models, which relies on CodeLLaMA~\cite{CodeLLaMA} as a base, also exhibits low tool hallucinations, suggesting that utilizing code-based approaches for tool learning is a viable direction.

\begin{table*}[!t]
\centering
\resizebox{\linewidth}{!}
{
\begin{tabular}{cc|cc|cc|c}       
\toprule
\multicolumn{2}{c|}{\textbf{ToolLLaMA-2-}}&\multicolumn{2}{c|}{\textbf{NexusRaven-}} & \multicolumn{2}{c|}{\textbf{GPT-}} & \multirow{2}*{\textbf{RoTLLaMA}} \\
\multirow{1}{0.15\linewidth}{\centering \textbf{7B-v1}} & \multirow{1}{0.15\linewidth}{\centering \textbf{7B-v2}} & \multirow{1}{0.15\linewidth}{\centering \textbf{13B-v1}} & \multirow{1}{0.15\linewidth}{\centering \textbf{13B-v2}} & \multirow{1}{0.2\linewidth}{\centering \textbf{3.5-turbo}} & \multirow{1}{0.1\linewidth}{\centering \textbf{4}} & \\
\midrule
53 & 65 & 6 & 0 & 50 & 23 & 3 \\
\bottomrule
\end{tabular}
}
\caption{The number of tool hallucinations for each LLM in all environments.}
\label{tab:hallucinations}
% % \vspace{-4mm}
\end{table*}

\paragraph{Performance of RoTToolLLaMA}
To confirm the robustness of our method for enhancing established tool-oriented LLMs, we proceed to fine-tune ToolLLaMA-2-7B using our generated trajectories and obtain RoTToolLLaMA. The corresponding results presented in Table~\ref{tab:RoTToolLLaMA} illustrate that our method's fine-tuning significantly enhances the model's tool learning capability across all stages, while also bolstering its overall robustness.
For instance, across the three stages, our method demonstrates performance extremes of 12.33/13.33/9.53 in various environments, compared to ToolLLaMA-2-7B-v2's 26.67/16.67/10.95. This further underscores the efficacy of our proposed approach.

\begin{table*}[!t]
\centering
% \resizebox{\linewidth}{!}
{
\begin{tabular}{l| ccccc} 
\toprule
\textbf{Level} & \textbf{Clean} & \textbf{Slight} & \textbf{Medium} & \textbf{Heavy} & \textbf{Union} \\
\midrule
$\mathbf{s_{TS}}$ & 69.52& 69.05& 70.95& 64.76& 56.19
\\
$\mathbf{s_{PI}}$ & 52.38& 45.24& 50.95& 40.95& 39.05
\\
$\mathbf{s_{CF}}$ & 38.10& 32.38& 34.76& 31.43& 28.57
\\
\bottomrule
\end{tabular}
}
\caption{The score in different stages (\%) of RoTToolLLaMA
in various Environments.}
\label{tab:RoTToolLLaMA}
\end{table*}

\section{Prompt Template for Inference}
\label{sec:prompt}

In the context of inference, both the ToolLLaMA-2-7B family of models and the GPT family of models utilize the same prompt (See Table~\ref{tab:prompt-infer}), whereas NexusRaven-13B-v1 and NexusRaven-13B-v2 employ distinct prompts (See Table~\ref{tab:prompt-neu} and Table~\ref{tab:prompt-neu-v2}).

\begin{table*}[!t]
    \centering
    \resizebox{\linewidth}{!}{
    \begin{tabular}{p{\linewidth}}
    \toprule
   \rowcolor{gray!10} \multicolumn{1}{c}{\textit{System}} \\
   You are an expert in using tools to handle real-time queries from users.\\
    First I will give you the task description, and your task start.\\
    At each step, your task is to give your thought to analyze the current state, decide the next step, with a function call to actually execute your step.\\
    After the call, you will get the call result, and you are now in a new state.\\
    Then you will analyze your status now, then decide what to do next...\\
    After many (Thought-call) pairs, you finally perform the task, then you can give your final answer.\\\\
    Desired format:\\
    Thought: $\langle$ The thought$\rangle$\\
    Action: $\langle$ The tool you decide to use$\rangle$\\
    Action Input: $\langle$ The parameters for the tool$\rangle$\\\\
    Remember:\\
    1. You should ALWAYS think about what to do, but all the thought is short, at most in 3 sentences.\\
    2. The action to take should be one of the given tools below.\\
    3. The ``Action Input'' needs to provide a dict similar to \{parameter\_1: value\_1, parameter\_2: value\_2\} to call action.\\
    4. Always use the ``finish'' tool upon task completion. The final answer should be comprehensive enough for the user. If the task is unmanageable, use the ``finish'' tool and respond with ``I cannot handle the task.''\\\\
    Task description: You should use tools to help handle the real time user queries. Specifically, you have access of the following tools:\\
    \{Tool Document\}\\\\
    Let's Begin!\\ \midrule%\hdashrule[0.5ex]{\linewidth}{1pt}{3mm}
    \rowcolor{gray!10} \multicolumn{1}{c}{\textit{User}} \\
    \{Query\}\\
    Begin!\\ %\hdashrule[0.5ex]{\linewidth}{1pt}{3mm}
    % Assistant:\\
    \bottomrule
    \end{tabular}
    }
    \caption{The prompt used for ToolLLaMA-2-7B family of models and GPT
    family of models, where ``\{Tool Document\}'' represents the tool documentation given to LLMs and ``\{Query\}'' represents the query given by the user.}
    \label{tab:prompt-infer}
\end{table*}

\begin{table*}[!t]
    \centering
    \resizebox{\linewidth}{!}{
    \begin{tabular}{p{\linewidth}}
    \toprule
    \rowcolor{gray!10} \multicolumn{1}{c}{\textit{User}} \\
    % <human>:\\
    % OPTION:\\
    % <func\_start>def \{func\_name\}(\{param\_name\} : \{param\_type\}, \{param\_name\} : \{param\_type\}) <func\_end>\\
    % <docstring\_start>\\
    % ```\\
    % \{func\_description\}\\\\
    % Args:\\
    % \{param\_name\} (\{param\_type\}) : \{param\_description\}\\
    % \{param\_name\} (\{param\_type\}) : \{param\_description\}\\
    % '''\\
    % <docstring\_end>\\\\
    % OPTION:\\
    % ...\\\\
    \{Tool Document\}\\\\
    User Query: Question: \{Query\}\\\\
    Please pick a function from the above options that best answers the user query and fill in the appropriate arguments.
    % <human\_end>
    \\
    \bottomrule
    \end{tabular}
    }
    \caption{The prompt used for NexusRaven-13B-v1, where ``\{Tool Document\}'' represents the tool documentation given to LLMs
    % where ``\{func\_name\}''/ ``\{func\_description\}'' represent the name/description of a tool given to the model, the name/type/description of a parameter owned by the tool are represented as ``\{param\_name\}''/ ``\{param\_type\}''/ ``\{param\_description\}'' respectively 
    and ``\{Query\}'' represents the query given by the user.}
    \label{tab:prompt-neu}
\end{table*}

\begin{table*}[!t]
    \centering
    \resizebox{\linewidth}{!}{
    \begin{tabular}{p{\linewidth}}
    \toprule
    \rowcolor{gray!10} \multicolumn{1}{c}{\textit{User}} \\
    % Function:\\
    % def \{func\_name\}(\{param\_name\} : \{param\_type\}, \{param\_name\} : \{param\_type\}):\\
    % ```\\
    % \{func\_description\}\\\\
    % Args:\\
    % \{param\_name\}: \{param\_description\}\\
    % \{param\_name\} (Optional): \{param\_description\}\\
    % '''\\\\
    % Function:\\
    % ...\\ \midrule
    % \rowcolor{gray!10} \multicolumn{1}{c}{\textit{User}} \\
    \{Tool Document\}\\\\
    User Query: \{Query\}\\
    \bottomrule
    \end{tabular}
    }
    \caption{The prompt used for NexusRaven-13B-v2, where ``\{Tool Document\}'' represents the tool documentation given to LLMs
    % where ``\{func\_name\}''/ ``\{func\_description\}'' represent the name/description of a tool given to the model, the name/type/description of a parameter owned by the tool are represented as ``\{param\_name\}''/ ``\{param\_type\}''/ ``\{param\_description\}'' respectively 
    and ``\{Query\}'' represents the query given by the user.}
    \label{tab:prompt-neu-v2}
\end{table*}

\section{Prompt Template for Query Expansion}
\label{sec:prompt-query}
We use GPT-4 for query expansion based on prompt in Table~\ref{tab:prompt-query}.

\begin{table*}[!t]
    \centering
    \resizebox{\linewidth}{!}
    {
    \begin{tabular}{p{\linewidth}}
    \toprule
    \rowcolor{gray!10} \multicolumn{1}{c}{\textit{System}} \\
    As an expert, your assignment is to utilize the comprehensive documentation of various tools to develop a series of problem scenarios that these tools can resolve. Ideally, each scenario should necessitate the sequential use of multiple tools for its resolution.\\
    \\
Remember:\\
1. The tools employed to address a problem should be a subset of the tools detailed in the provided documentation; ideally, each problem should require the use of more than one tool.\\
2. The parameter values needed by each tool can either be directly extracted from the query or obtained by invoking the specified other tool.\\
3. The problem scenario should be expressed in a way that is understandable to humans, while also showcasing the diverse functions of the provided tools and their interrelationships.\\
\\
Here is the documentation of various tools: \{Tool Document\}\\
\midrule
\rowcolor{gray!10} \multicolumn{1}{c}{\textit{User}} \\
Please generate 12 diverse queries according to the documentation.\\
\\
Examples:\\
\{Examples\}\\
\bottomrule
    \end{tabular}
    }
    \caption{The prompt for query expansion, where ``\{Tool Document\}'' represents the tool documentation given to LLMs and ``\{Examples\}'' represents the examples for LLMs.}
    \label{tab:prompt-query}
\end{table*}

\section{Prompt Template for Trajectory Generation}
\label{prompt-trajectory}
We use GPT-4 for trajectory generation based on prompt in Table~\ref{tab:prompt-trajectory}.

\begin{table*}[!t]
    \centering
    \resizebox{\linewidth}{!}
    {
    \begin{tabular}{p{\linewidth}}
    \toprule
    \rowcolor{gray!10} \multicolumn{1}{c}{\textit{System}} \\
    You are an expert in using tools to handle real-time queries from users.\\
At each step, your task is to give your thought to analyze the current state, decide the next step, with a function call to actually execute your step.\\
After the call, you will get the call result, and you are now in a new state.\\
Then you will analyze your status now, then decide what to do next...\\
After a series of these thought-action pairs, you will complete the task and provide the final answer.\\
\\
Remember:\\
1. You must ALWAYS select a specific function to execute your idea at each step.\\
2. Before calling any function, you should ALWAYS give your thought, but limit it to a maximum of three sentences.\\
3. ALWAYS use the ``finish'' tool upon task completion. The final answer should be comprehensive enough for the user. If the task is unmanageable, use the ``finish'' tool and respond with ``I cannot handle the task''.\\
\\
Let's begin!\\
\midrule
\rowcolor{gray!10} \multicolumn{1}{c}{\textit{User}} \\
\{Query\}\\
Begin!\\
\bottomrule
    \end{tabular}
    }
    \caption{The prompt for trajectory generation, where ``\{Query\}'' represents the query given by the user.}
    \label{tab:prompt-trajectory}
\end{table*}

% \section{Example Appendix}
% \label{sec:appendix}

% This is an appendix.

\end{document}